\documentclass[10pt, a4paper]{article}

\usepackage[final]{lrec-coling2024} 

\newcommand{\tabH}{\rule{0pt}{2ex}}
\newcommand{\bhline}{\noalign{\hrule height 1.2pt}}
\newcommand{\nli}{$\mbox{en-NLI}$}
\usepackage{multirow}
\usepackage{amsmath}
\usepackage{graphicx}
\usepackage{siunitx}
\usepackage{url}
\usepackage{amssymb}
\usepackage{xcolor,colortbl}
\usepackage{cleveref}
\crefname{table}{Table}{Tables}
\crefname{figure}{Figure}{Figures}
\crefname{section}{Section}{Sections}
\crefname{appendix}{Appendix}{Appendix}
\PassOptionsToPackage{hyphens}{url}

\title{ Multilingual Sentence-T5: \\Scalable Sentence Encoders for Multilingual Applications}

\name{
  Chihiro Yano${}^{1,2}$ Akihiko Fukuchi${}^{2}$ Shoko Fukasawa${}^{2}$\\
{\bf \large Hideyuki Tachibana${}^{2}$ Yotaro Watanabe${}^{2}$}}

\address{${}^{1}$Graduate School of Informatics, Nagoya University
${}^{2}$PKSHA Technology Inc.\\
yano.chihiro.j3@s.mail.nagoya-u.ac.jp\\
\{akihiko\_fukuchi,shoko\_fukasawa,h\_tachibana,y\_watanabe\}@pkshatech.com}

\abstract{
Prior work on multilingual sentence embedding has demonstrated that the efficient use of natural language inference (NLI) data to build high-performance models can outperform conventional methods. However, the potential benefits from the recent ``exponential'' growth of language models with billions of parameters have not yet been fully explored. 
In this paper, we introduce Multilingual Sentence T5 (m-ST5), as a larger model of NLI-based multilingual sentence embedding, by extending Sentence T5, an existing monolingual model. 
By employing the low-rank adaptation (LoRA) technique, we have achieved a successful scaling of the model's size to 5.7 billion parameters. 
We conducted experiments to evaluate the performance of sentence embedding and verified that the method outperforms the NLI-based prior approach. 
Furthermore, we also have confirmed a positive correlation between the size of the model and its performance.
It was particularly noteworthy that languages with fewer resources or those with less linguistic similarity to English benefited more from the parameter increase.
Our model is available at \url{https://huggingface.co/pkshatech/m-ST5}.
 \\ \newline  }

\begin{document}
\maketitleabstract

\section{Introduction}
Sentence embedding is a versatile and fundamental technique of NLP and has been studied extensively \citep{SkipThought,QuickThought,SBERT,ConSERT, COCO-LM, carlsson2021semantic, kim-etal-2021-self,SGPT}. In particular, the recently proposed SimCSE \citep{SimCSE}, a simple and data-efficient method based on contrastive fine-tuning of existing pre-trained text encoders such as BERT, greatly advanced the frontier and attracted much attention.
This technique can be naturally used with other kinds of model architectures. 
For example, in their Sentence T5, \citet{ST5} adopted T5, an encoder-decoder model.
\begin{figure}[t!]
\centering
\includegraphics[width=1\linewidth]{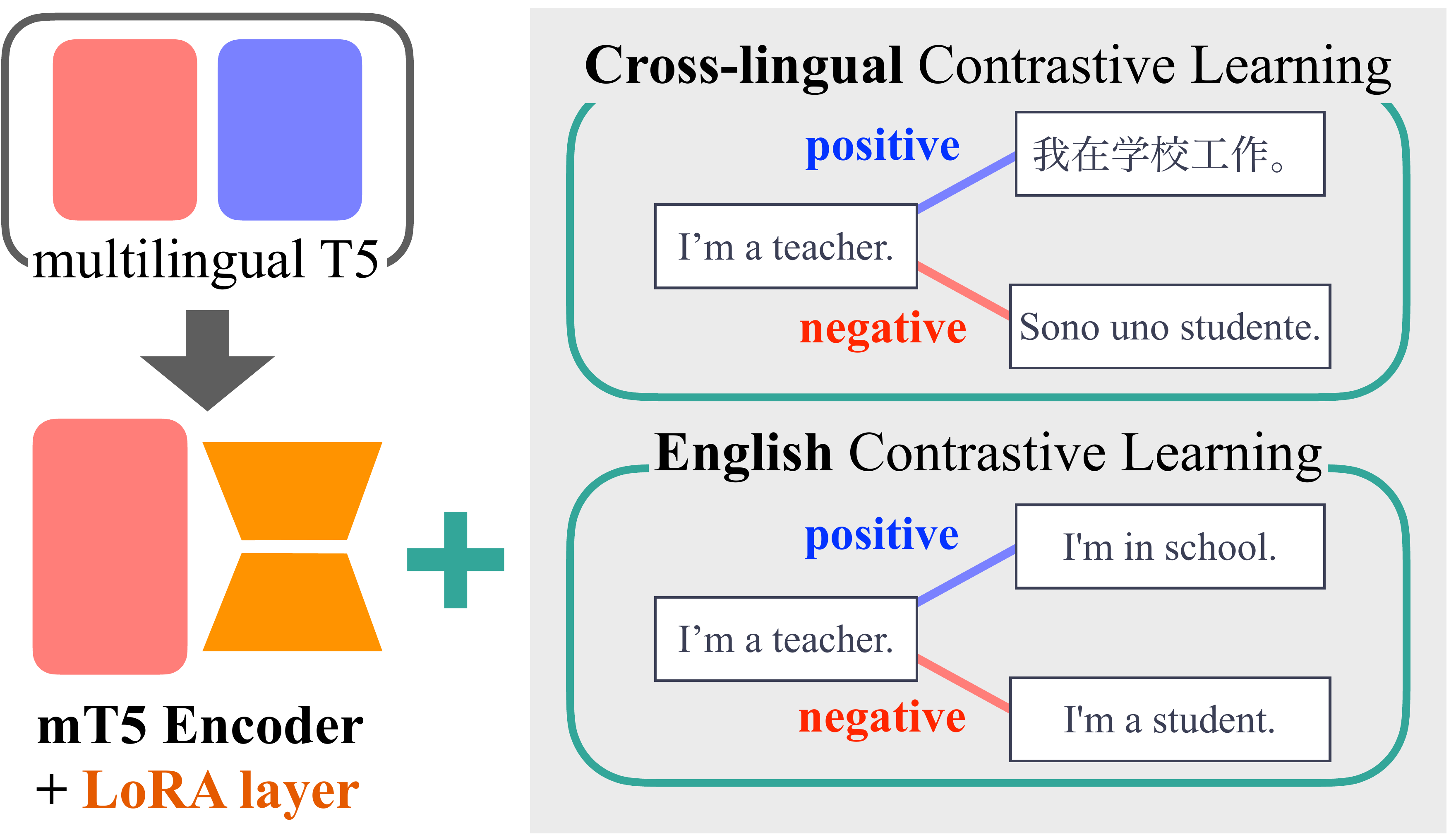}
\caption{\mbox{Concept diagram of m-ST5.}}
\label{fig:over-view}
\vspace{2ex}
\includegraphics[width=1\linewidth]{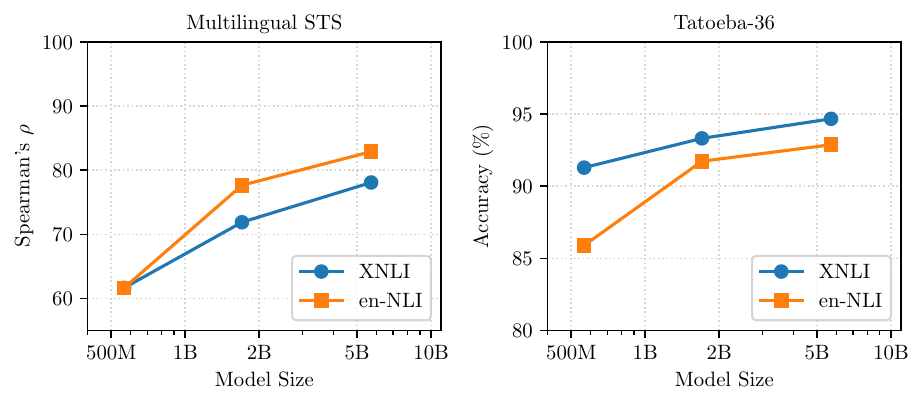}
\caption{Comparison of model size and performance of the proposed method.}
\label{fig:scaling law}
\end{figure}

Multilingual sentence embedding, which projects sentences from diverse languages into a shared semantic space, is an important extension of this problem, and many techniques have been proposed (\cref{sec:related_work}).
Of these, we particularly focus on a multilingual extension of SimCSE, namely mSimCSE \citep{mSimCSE}, because of its data efficiency. In particular, even though the fine-tuning requires just a natural language inference (NLI) dataset consisting of around 2 million sentences, 
it showed comparable results with supervised techniques based on larger parallel corpora.

Now, the natural question that arises here is whether such a learning strategy scales to larger models with billions of parameters.
To answer this question, this paper examines the performance of a fine-tuned model based on mT5 \citep{mT5}, a larger model than XLM-RoBERTa \citep{XLM-R}, which is the base model of mSimCSE. 
We extend the existing large-scale monolingual model Sentence T5 to multilingual scenarios. 

The proposed method performed well on some benchmarks, including cross-lingual STS (XSTS) \citep{XSTS} and sentence retrieval \citep{LASER, bucc}. 
Besides, it outperformed a monolingual model in Japanese, a language far distant from the ones used for training.
We have further confirmed a positive correlation between the model size and its performance, as shown in \cref{fig:scaling law}, which is often referred to as the scaling law and found in other language models \citep{ST5, scaling}.
The observation that the scaling law holds for multilingual sentence embeddings suggests that the constraint of insufficient amount of training data in low-resource languages may be alleviated by using large-scale pre-trained models.

\newcommand{\reswang}{${}^*$ }
\newcommand{\resfeng}{${}^\dagger$ }
\newcommand{\resreim}{${}^\ddagger$ } 
\newcommand{\resph}{$\phantom{{}^*}$ }
\begin{table*}[t!]

\centering
\renewcommand{\tabcolsep}{0.5ex}

{\small
\begin{tabular}{@{}ccccccccccc@{}} \bhline
\tabH  &Train& Fine Tuning & \multicolumn{2}{c}{Tatoeba tasks (Accuracy)} &BUCC ($F_1$) & \multicolumn{5}{c}{XSTS tasks (Spearman's $\rho$)} 
\\      Model            & method &  Data  & Tatoeba-14 & Tatoeba-36 & avg.  & ar-ar & ar-en & es-es & es-en & tr-en \\ \bhline
\multicolumn{10}{c}{\tabH  \textbf{Contrastive Learning}} \\ \hline
\tabH \multirow{2}{*}{mSimCSE}  & \multirow{2}{*}{Full FT}& XNLI      & 93.2\reswang & 91.4\reswang & 95.2\reswang & 79.4\reswang & 72.1\reswang & 85.3\reswang & 77.8\reswang & 74.2\reswang \\ 
                                && \nli{}      & 89.9\reswang & 87.7\reswang & 93.6\reswang & 81.6\reswang & 71.5\reswang & 87.5\reswang & 79.6\reswang & 71.1\reswang\\ \hline
\tabH \multirow{4}{*}{m-ST5}  & LoRA& XNLI      &96.3\resph & 94.7\resph & 97.6\resph & 76.2\resph & 78.6\resph & 84.4\resph & 76.2\resph & 75.1\resph\\ 
                                &(query+value) & \nli{}      & 93.9\resph & 92.9\resph & 96.6\resph & 83.2\resph & 79.5\resph & 87.7\resph & 84.9\resph & 79.2\resph\\ 
 \cline{2-11}
\tabH & LoRA & XNLI & \textbf{96.5}\resph & 94.8\resph & \textbf{97.7}\resph & 77.3\resph & 77.8\resph & 85.0\resph &	77.7\resph & 75.0\resph \\ 
& (all-linear)& \nli{} & 94.0\resph & 93.1\resph & 96.7\resph & \textbf{84.5}\resph & \textbf{82.9}\resph & \textbf{89.2}\resph & \textbf{86.3}\resph & \textbf{79.7}\resph \\

\hline
\multicolumn{10}{c}{\tabH \textbf{Fully Supervised}} \\ \hline
\tabH LASER               &-      & -         & 95.3\resfeng & 84.4\resfeng & 93.0\resreim & 68.9\resreim & 66.5\resreim & 79.7\resreim & 57.9\resreim & 72.0\resreim\\
LaBSE                   &-        & -         & 95.3\resfeng & \textbf{95.0}\resfeng & 93.5\resreim & 69.1\resreim  & 74.5\resreim  & 80.8\resreim  & 65.5\resreim  & 72.0\resreim\\ 
\bhline
\end{tabular}
}
\caption{Evaluation results using Tatoeba, BUCC, and XSTS. 
Each score is the average of the performance over three trials with different random seeds.
Scores with \reswang, \resfeng and \resreim were excerpted from 
\citep{mSimCSE}, \citep{LaBSE}, and \citep{Sentence_Transformers}, respectively.
}
\label{tab:main}
\vspace{2.5ex}
\centering
\small
\begin{tabular}{ccc cccccccccc} \bhline
\renewcommand{\tabcolsep}{2ex}

\tabH Model & Train method & FT Data & hi & fr & de & af & te & tl & ga & ka & am & sw \\ \bhline
\multicolumn{12}{c}{\tabH\textbf{ Contrastive Learning}} \\ \hline
\tabH \multirow{2}{*}{mSimCSE}& \multirow{2}{*}{Full FT} & XNLI & 96.2 & 94.8 & 98.8 & 90.6 & 96.2 & 80.9 & 65.1 & 92.4 & 82.4 & 67.8 \\
& & \nli{} & 94.4 & 93.9 & 98.6 & 85.6 & 92.9 & 70.0 & 54.8 & 89.2 & 79.5 & 42.1\\ \hline
\tabH \multirow{4}{*}{m-ST5} & LoRA & XNLI & \textbf{98.0} & 96.2 & \textbf{99.7} & 95.6 & 98.2 & 94.3 & 83.0 & 95.4 & 93.1 & 91.2 \\
&(query+value)&\nli{} & 97.5 & 95.7 & 99.4 & 94.5 & 97.3 & 93.1 & 81.6 & 93.3 & 90.7 & 68.5\\ \cline{2-13}
\tabH & LoRA & XNLI & 97.8 & \textbf{96.4} & 99.6 & 95.6 & 97.9 & 94.1 & 84.3 & 95.6 & \textbf{94.3} & \textbf{91.5}\\
&(all-linear) & \nli{} &97.6 & 95.7 & 99.3 & 94.5 & 97.2 & 93.5 & 82.4 & 93.9 & 91.7 & 68.8\\
\hline
\multicolumn{12}{c}{\tabH \textbf{Fully Supervised}} \\
\hline
\tabH  LASER &-& - & 94.7 & 95.7 & 99.0 & 89.4 & 79.7 & - & \phantom{0}5.2 & 35.9 & 42.0 & 42.4 \\
\tabH  LaBSE &-& - & 97.8 & 96.0 & 99.4 & \textbf{97.4} & \textbf{98.3} & \textbf{97.4} & \textbf{95.0} & \textbf{95.9} & 94.0 & 88.5\\
\bhline
\end{tabular}
\caption{Accuracy of Tatoeba 
retrieval task. Target languages are the same as in \citep{mSimCSE}.}
\label{tab:tatoeba}
\renewcommand{\tabcolsep}{1ex}
\end{table*}

\section{Related Work}\label{sec:related_work}
Recently, multilingual models \citep{BERT,XLM, XLM-R, HICTL,XLM-E, mT5} have been intensively studied as they have language transferability, the ability to adapt to a new language in a few or zero shots.
It should also be noted that the transferability varies by task and language pair and is not always effective, especially between distant languages \citep{how_mBERT,lauscher-etal-2020-zero}.

As an important branch of multilingual NLP, various techniques for multilingual sentence embedding have been studied.
A major challenge in this field is how to acquire semantic proximity between sentences in different languages.
A natural approach would be the use of parallel corpora \citep{LASER,muse, LaBSE},
though this approach is data-hungry, and it is costly to maintain such corpora.
Some techniques have been proposed that can alleviate such problems, such as a distillation-based approach \citep{Sentence_Transformers}, introduction of adversarial training strategy to reduce language identifier information from semantic vectors \citep{chen-etal-2019-multi-source, keung-etal-2019-adversarial},
the use of word alignment between parallel sentences \citep{Cao2020Multilingual}, and so on.

In this paper, we particularly focus on the NLI-based contrastive training strategy, specifically the multilingual extension of SimCSE \citep{SimCSE}, namely mSimCSE \citep{mSimCSE}. 
It was shown that mSimCSE could acquire inter-language alignments without explicit parallel corpora, and even monolingual NLI corpora could yield good fine-tuning results.

\section{Proposed Method: m-ST5} 
In this paper, we propose the \textit{Multilingual Sentence T5} (m-ST5) as a new extension of Sentence T5 \citep{ST5}.
Similarly to ST5, our method is based on fine-tuning of a pre-trained T5 model \citep{T5},
which is one of the most popular encoder-decoder language models.
However, since we are interested in multilingual sentence embedding, we need to use a multilingual model as our baseline. We then used Multilingual T5 (mT5) \citep{mT5}, 
which was pre-trained on mC4, a large-scale multilingual corpus covering 101 languages.
To build a sentence embedding model, only the encoder part of the enc-dec model is needed, as in Sentence T5. 
For example, the encoder module (5.7B params) out of the pre-trained mT5-xxl (13B params) is extracted.
The encoder converts a sentence into token-wise embedding, and these token representations are averaged together to produce a sentence embedding.

Naturally, the vector obtained this way is not sufficient for sentence embedding in quality, and fine-tuning of the encoder is required.
To this end, following mSimCSE \citep{mSimCSE}, we trained \mbox{m-ST5} in a contrastive manner using the NLI dataset
for a task to predict whether a given hypothesis sentence is an \textit{entailment}, a \textit{contradiction}, or \textit{neutral} to another premise sentence.
Specifically, m-ST5 is trained to minimize the distances between positive pairs (entailment) and maximize the distances between negative ones (contradiction). Furthermore, unrelated in-batch sentences are also incorporated as negative samples because 
such a trick promotes the uniformity of the semantic space \citep{SimCSE}.

Additionally, in multilingual learning scenarios using cross-lingual NLI data (XNLI) \citep{XNLI}, it should also be taken into account which languages the positive and negative samples are drawn from.
In this study, following mSimCSE, we draw triplets, each of which consists of a premise and two hypotheses (entailment and contradiction) from different languages, as shown in \cref{fig:over-view}.

\section{Experiment}
\subsection{Cross-lingual Experiments}
We first evaluated the quality of sentence embedding by sentence retrieval tasks (Tatoeba, \citealp{LASER}; and BUCC, \citealp{bucc}) and cross-lingual STS task~(XSTS, \citealp{XSTS}).
Details of the evaluation task are provided in \cref{sec:appendix-eval}.
For all tasks, cosine similarity was used as the measure of similarity.
We compared the performance with the following methods: mSimCSE \citep{mSimCSE}, LASER \citep{LASER}, and LaBSE \citep{LaBSE}.
Of these, LASER and LaBSE were trained in a fully supervised manner.

In order to feasibly train our models on a single A100 GPU with 80GB of VRAM, we used the technique of LoRA \citep{LoRA}, which enables training of very large models with limited computational resources.
In this paper, we examined two LoRA conditions.
One is to apply the LoRA technique only to the query and value matrices, following the original LoRA paper \citep{LoRA}, which reported that fine-tuning only the query and value matrices is effective.
The other is to apply LoRA to all linear layers, following 
the QLoRA paper \citep{QLoRA} which showed that the highest performance is obtained by fine-tuning all linear layers. 
In both cases, the rank of the matrices was $r=8$,
and the batch size was 128 in our experiment.
Other details of the experimental setup are in \cref{sec:appendix-train}.
The training data was chosen from the following NLI datasets: \nli{} $=$ SNLI \citep{SNLI} $+$ MNLI \citep{N18-1101},
and XNLI \citep{XNLI}. 
Details of the training data are in \cref{sec:appendix-train-data}.

\cref{tab:main} shows the evaluation results on the above three tasks.
We may observe that the proposed method (m-ST5) outperformed the existing mSimCSE. 
Notably, even monolingual (\nli{}) fine-tuning of m-ST5 outperformed multilingual (XNLI) fine-tuning of mSimCSE.
It is also observed that all-linear LoRA gave better results than query+value LoRA.
The difference was especially noticeable when \nli{} data was used for training.

We also found that it depends on the task which of XNLI/\nli{} fine-tuning data gave better results. 
Specifically, we have observed that cross-lingual training data (XNLI) was notably more effective in sentence retrieval tasks, while monolingual data (\nli{}) was more effective for the XSTS task.
It may reflect differences in the nature of the evaluation metrics, i.e.\ the sentence retrieval tasks are only concerned with the sentences with top relevance, while the STS task considers ranking that requires more elaborate knowledge on each concept.
Here, the XNLI-based learning prioritizes the alignment of different languages, resulting in sparse occurrences of each word and making it harder to acquire elaborated word knowledge.
In this respect, monolingual learning using \nli{}, with its opposite properties where the same word often appears both in positive and negative samples, would have been more effective in STS.

\cref{tab:tatoeba} further details the accuracy of 
\mbox{English-*} sentence retrieval in various languages.
While mSimCSE did not perform well for low-resource languages (e.g.\ ga) or phylogenetically distant languages from English (e.g.\ sw), the proposed method produced high scores of over 90\% for all of such languages except Irish (ga).
Overall, the performance has improved, and the results are approaching those of fully supervised methods.

\subsection{Comparison with Monolingual Models}
\label{sec:comparison-with-mono}
To evaluate the transferability of the proposed method, we conducted experiments on Japanese, Korean and Chinese languages which are phylogenetically very distant from English but are rich in evaluation resources and the population of potential users.
The performance of our method (m-ST5) on monolingual STS tasks was compared with these monolingual models, as well as the multilingual models.
\paragraph{Baseline Models}
As monolingual baselines, we used the Japanese BERT-large
\footnote{\url{https://huggingface.co/cl-tohoku/bert-large-japanese-v2}}, the Korean RoBERTa-large from KLUE \citep{klue}, 
and the Chinese RoBERTa-large  \citep{zho-roberta}.
We will call these models `ja-BERT',  `ko-RoBERTa' and `zh-RoBERTa' for simplicity.
As multilingual models, we used the LaBSE and mSimCSE.
The LaBSE model was from Hugging Face Hub\footnote{\url{https://huggingface.co/sentence-transformers/LaBSE}}, and the mSimCSE model was reproduced based on the original paper \citep{mSimCSE}.

\paragraph{Fine-tuning and Evaluation Data}
The training data used for fine-tuning were XLNI, \nli{}, and monolingual NLI datasets for each language. 
(Note that XNLI do not contain Japanese and Korean languages.)
For evaluation, STS data in each language, as well as English STS data (STS-B, \citealp{STSB}) were used. 
Note that the English STS is not the main focus of this section, but is for reference.

The Japanese monolingual NLI dataset was JSNLI \citep{JSNLI},
and the evaluation STS dataset was JSTS in JGLUE \citep{jglue}.
The Korean dataset was KorNLI/KorSTS \citep{kornli}. 
The Chinese NLI dataset was CMNLI in CLUE \citep{clue}, and the STS dataset was STS-B test set in \mbox{C-MTEB} \citep{CPackPR}.
We will refer to these fine-tuning and evaluation data as 
\{ja, ko, zh\}-\{NLI, STS\}, for simplicity.

\begin{table*}[t]
\centering
\begin{tabular}{@{}cccccccc@{}} \bhline
\tabH
Model                  &Train method  & FT Data  &en-STS&ja-STS & ko-STS & zh-STS & avg.\\ \bhline
\renewcommand{\arraystretch}{1.1}
\tabH
ja-BERT &\multirow{3}{*}{Full FT} & ja-NLI &-- & 83.6 & -- & -- &--\\
ko-RoBERTa && ko-NLI&-- & -- & \textbf{83.4} & --&-- \\ 
zh-RoBERTa && zh-NLI &--& -- & -- & 71.2&-- \\ 
\hline
\tabH \multirow{4}{*}{\vtop{\hbox{\strut m-ST5}\hbox{\strut (ours)}}} 
    &\multirow{2}{*}{\begin{tabular}{c}LoRA\\(query+ value)\end{tabular}}
    & XNLI &80.3& 81.4 & 73.4 & 73.0&77.3 \\
    && \nli{} &85.6& 82.7 & 77.2 & 77.3& 80.7\\
    \cline{2-8}
    \tabH
    &\multirow{2}{*}{\begin{tabular}{c}LoRA\\(all-linear)\end{tabular}}
    & XNLI &84.2& 82.1 &78.0 & 74.7&79.8 \\
    && \nli{} &\textbf{88.1}& \textbf{84.1} & 81.1 & 79.6& \textbf{83.2}\\
    \hline
\tabH \multirow{2}{*}{mSimCSE}  &\multirow{2}{*}{Full FT} & XNLI& 78.3& 79.0 & 72.9 & 69.6&75.0 \\ 
    & & \nli{} & 87.2& 81.2 & 80.1 & \textbf{80.4} &82.2\\ \bhline
\tabH LaBSE &-- & -- &74.1& 76.1 & 70.5 & 68.4&72.3 \\ \bhline
\end{tabular}
\caption{Comparison with monolingual models. 
`ja', `ko' and `zh' 
refer to Japanese, Korean and Chinese, respectively.
LaBSE was evaluated using a model published on Hugging Face Hub, and mSimCSE was evaluated using a model reproduced based on the original paper.
}
\label{tab:monosts}
\end{table*}
\paragraph{}

\cref{tab:monosts} shows the evaluation results of STS tasks in three languages mentioned.
The proposed model with the LoRA layer added only to query+value matrices showed inferior results to the existing multilingual model, mSimCSE.
Nevertheless, this problem was solved by adding the LoRA layers to all linear layers, and by increasing the number of trainable parameters \citep{QLoRA}. 

The average score of the results for these languages was higher than that of the existing multilingual models.
The performance for each language was as follows:
in Chinese, the proposed method outperforms the monolingual counterpart (zh-RoBERTa), and in Japanese, the performance of the proposed method is equivalent to that of the monolingual counterpart (ja-BERT).
In particular, even when the target language data was not used for training at all (i.e., only \nli{} was used), the performance of m-ST5 was comparable to these monolingual models. 

This could be attributed to two factors: The first would be that m-ST5 has high cross-lingual transferability, and the second would be the large size and high quality of the \nli{} dataset.

These results suggest that fine-tuning multilingual models for monolingual tasks is a promising option when pre-trained large monolingual models are not available.
Moreover, high performance could be achieved without using target language data during fine-tuning.

In Korean language, however, the opposite trend has been observed. The monolingual model trained on the monolingual corpus was significantly better than m-ST5, transfer-learned from a multilingual model.
In our observation, this could be attributed to the quality of tokenizers. In general, the tokenization of Korean language is not a straightforward task \citep{klue}, and the tokenizer of our base multilingual model does not seem to be sufficiently tuned in this respect. On the other hand, the tokenizer of ko-RoBERTa seems to have been carefully crafted.

\begin{table}[t!]
\centering
\small
\renewcommand{\tabcolsep}{0.7ex}
\begin{tabular}{ccccccc} \bhline
\tabH Model & FT Data & ar-ar & ar-en & es-es & es-en & tr-en \\ \bhline
\tabH mT5-large& XNLI & 62.0 & 58.2 & 77.6 & 56.8 & 53.8 \\
 (564M) & \nli{} & 68.8 & 49.1 & 82.1 & 59.5 & 48.8 \\ \hline
\tabH mT5-xl & XNLI & 71.8 & 71.7 & 81.7 & 68.0 & 66.3 \\
 (1.7B) & \nli{} & 78.2 & 73.9 & 87.3 & 76.8 & 72.1 \\ \hline
\tabH mT5-xxl & XNLI & 76.2	& 78.6 & 84.4 & 76.2 & 75.1 \\
(5.7B) & \nli{} &  \textbf{83.2} & \textbf{79.5} & \textbf{87.7} & \textbf{84.9} & \textbf{79.2} \\ 
\bhline
\end{tabular}
\caption{Comparisons of models' performance on SemEval 2017 STS shared task when scaling up model size.}
\label{tab:model_size}
\renewcommand{\arraystretch}{1.0}
\end{table}

\subsection{Scaling Law}

It has been suggested that the performance of language models scale with the increase of the model size. 
In this section, we investigate whether this is the case for our approach as well as Sentence T5 \citep{ST5}.
We compared the performance of three pre-trained mT5 models of different sizes (564M, 1.7B, and 5.7B) when fine-tuned query and value matrices using XNLI or \nli{}.

\cref{fig:scaling law} shows the scaling law of the performance of Spearman's $\rho$ on XSTS and the accuracy of multilingual sentence retrieval on Tatoeba-36. 
\cref{tab:model_size} details the performance on XSTS for various language pairs. 
In this table,
a trend could be observed that languages far from English (i.e., ar and tr) tended to benefit more from the increase in model size.
Particularly interesting in this table is the fact that monolingual fine-tuning becomes more effective as the model size is scaled up. This result suggests that cross-lingual transferability emerges when the model becomes larger.

\section{Conclusion}
In this paper, we proposed the Multilingual Sentence T5 (m-ST5), an extension of Sentence T5 to multilingual. 
m-ST5 demonstrated excellent performance in multilingual tasks such as cross-lingual sentence retrieval and cross-lingual STS. It also performed well in a monolingual task, demonstrating the effectiveness of the proposed model in low-resource languages where no large-scale, high-performance model exists. 
Furthermore, we investigated the correlation between the size of the model's parameters and changes in performance, confirming that performance changes follow the scaling laws and that performance improvements are particularly notable in low-resource languages. 
Note that we are planning to release the trained model of the proposed m-ST5. 

\section*{Ethics Statement}
Since this method is an embedding model and does not generate language, the risk of generating harmful sentences would not need to be considered.
On the other hand, since the biases contained in the training data are incorporated as is, the vectors generated may implicitly contain such biases, and the possibility of serious discriminatory results in some applications cannot be denied. In actual applications, maximum measures are needed to prevent such disadvantages.

\section{Bibliographical References}
\bibliographystyle{lrec-coling2024-natbib}
\bibliography{ref}

\section{Evaluate task detail}
\label{sec:appendix-eval}
\paragraph{Tatoeba-\{14 and 36\}:}
A community-based corpus of English sentences and their translations into more than 400 languages, from which \citet{LASER} generated a dataset for multilingual NLP tasks. 
We conduct evaluations on sentence retrieval tasks with pairs of English sentences and sentences in other languages.
There are two settings: one with 14 languages and another with 36 languages.
\paragraph{BUCC:} It is a bitext mining task to predict translated sentences from a collection of sentences in two languages. It consists of English and one of the 4 languages (German, French, Russian and Chinese) \citep{bucc}.
Following XTREME \citep{XTREME}, 
we regarded sentence pairs whose similarity exceeded the pre-defined threshold as translations of each other, and the results were evaluated using F-measure.

\paragraph{XSTS:} The cross-lingual semantic textual similarity (XSTS) \citep{XSTS}, which is a multilingual extension of the vanilla STS that evaluates the correlation of the ranking of semantic similarity with human judgement.
The sentence pairs of the dataset are either in the same language or different languages.

\section{Training detail}
\label{sec:appendix-train}
\paragraph{Batch size and number of epochs:}
In our experiment, the batch size was 128.
In the preliminary experiments, batch sizes of 64, 128, and 256 were considered, but no significant differences were found.
Also, the number of epochs was set to 1.

\paragraph{Learning rate}
We used AdamW as the optimizer \citep{AdamW}.
We investigated the learning rate at \num{e-5}, \num{5e-5}, \num{e-4}, \num{5e-4}, and used the one that showed the best performance on the STS Benchmark \citep{XSTS} dev set. 

\paragraph{LoRA configuration:}
The rank of the LoRA adaptation matrix was $r=8$ and the weight was $\alpha=32$.
We tried ranks $r = 4, 8,$ and $16,$ but did not observe significant differences in performance.
The details of the differences are shown in \cref{tab:LoRA_r}.

\paragraph{Model size and training data volume} 
\label{sec:appendix-model}

\cref{tab:model_detail} compares the model size and training data volume.

\begin{table}[!t]
\centering
\small
\renewcommand{\tabcolsep}{0.9ex}
\renewcommand{\arraystretch}{1.1}
\begin{tabular}{ccccccc}
\bhline
\tabH \multirow{2}{*}{Data} & \multirow{2}{*}{$r$}  & \multirow{2}{*}{lr} & XSTS & tatoeba & tatoeba & BUCC \\ 
& & &avg. &14 &36 &avg. \\ \bhline
\tabH \multirow{3}{*}{XNLI}   & 4  & 5e-5 & \textbf{78.1} & 96.2       & 94.6       & \textbf{97.6} \\
       & 8  & 5e-5 & \textbf{78.1} & \textbf{96.3}       & \textbf{94.7}      & \textbf{97.6} \\
       & 16 & 1e-5 & 76.3 & 96.1       & 94.6       & 97.4 \\ \hline
\tabH \multirow{3}{*}{\nli}    & 4  & 5e-4 & \textbf{83.0} & \textbf{94.0}       & \textbf{92.9}       & 96.5 \\
       & 8  & 5e-4 & 82.9 & 93.9       & \textbf{92.9}       & 96.6 \\
       & 16 & 5e-4 & 82.7 & 93.8       & 92.6       & \textbf{96.8} \\ \bhline
\end{tabular}
\caption{Performance when changing the rank of the LoRA adaptation matrix.}
\label{tab:LoRA_r}
\renewcommand{\arraystretch}{1.0}
\end{table}

\begin{table}[!t]
    \centering
\small
\renewcommand{\tabcolsep}{0.9ex}
    \begin{tabular}{ccc} \bhline
        \tabH Model & Size &Training Data  \\
        \bhline
        \tabH LASER & 0.2B & 200M pairs \\ \hline
        \tabH LaBSE &1.8B & 17B sents + 6B pairs  \\ \hline
        \tabH mSimCSE$_{\text{XNLI}}$ & 0.3B&2.5TB data + 2M pairs   \\ \hline
        \tabH m-ST5$_{\text{XNLI}}$ & 5.7B&6.3T tokens + 2M pairs  \\ 
        \tabH m-ST5$_{\text{\nli{}}}$ & 5.7B&6.3T tokens + 0.2M pairs  \\ 
        \bhline
    \end{tabular}
    \caption{Size and data datails of the models used in the experiment.}
    \label{tab:model_detail}
\end{table}

\section{Training data detail}
\label{sec:appendix-train-data}
In contrastive learning with NLI dataset, premise-hypothesis pairs that share the same premise are concatenated to create a triplet consisting of a premise and two hypotheses (entailment and contradiction).

\paragraph{\nli{}:}
This training dataset is the concatenation of Stanford NLI \citep{SNLI} and MultiNLI \citep{N18-1101}. Both datasets contain only English texts.
The ready-to-use dataset consisting of triplets made from these datasets can be downloaded from the official repository of SimCSE\footnote{\url{https://github.com/princeton-nlp/SimCSE}}\footnote{\url{https://huggingface.co/datasets/princeton-nlp/datasets-for-simcse}}. This dataset consists of  275,601 triplets.

\paragraph{XNLI:}
The XNLI dataset is a crowd-sourced translation of MultiNLI dataset into 15 languages. 
The number of triplets included in the training data was 1,963,485.

\paragraph{JSNLI:}
The JSNLI dataset is a machine translation of the SNLI dataset into Japanese \citep{JSNLI}. 
The train set contains around 533k premise-hypothesis pairs with labels.
By the pre-processsing described above, we obtained 176,309 triplets.

\section{Languages used in the experiment}
\label{sec:appendix-lang}

\cref{table:list-of-languages} shows the list of languages used in the experiments in this paper.
All the languages displayed in this table are included in mC4, the pre-training data for mT5 \citep{mT5}. Note that in the mC4 specification document, Hebrew and Tagalog (Filipino) are denoted as `iw' and `fil', respectively.

\newcolumntype{P}{>{\centering\arraybackslash}p{0.38cm}}
\newcommand{\yes}{\cellcolor{teal!25}\checkmark}
\begin{table*}[!t]
    \centering
    \scriptsize
    \begin{tabular}{@{}lllPPPPPPPPPP}
        \bhline
        code & language & family &
        \multicolumn{3}{c}{Tatoeba} 
        & \!\!\!\!BUCC 
        & \!\!\!XSTS 
        & \!\!XNLI 
        & \!\!\!\nli{} 
        & \!\!\!\!JSNLI 
        & \!\!\!\!KorNLI 
        & \!\!\!\!CMNLI\\
        & & & 
        14 & 36 & \!\!\!\!\!\cref{tab:tatoeba} 
        & & & & &&\\
        \bhline
        en & English 
            & IE / Germanic
            & \yes & \yes &\yes & \yes & \yes & \yes & \yes & &\\
        \hline
        af & Afrikaans 
            & IE / Germanic
            & & \yes & \yes & &  & & &&&\\
        am & Amharic 
            & Semitic
            & &  & \yes & & & & &&&\\
        ar & Arabic 
            & Semitic 
            & \yes  & \yes &  & & \yes & \yes & &&& \\
        bg & Bulgarian  
            & IE / Balto-Slavic
            & \yes & \yes &  & & & \yes & &&&\\
        bn & Bengali  
            &IE / Indo-Iranian
            & & \yes &  & & & & &&&\\
        de & German  
            &IE / Germanic
            & \yes & \yes & \yes & \yes & \yes & \yes & &&&\\
        el & Greek  
            &IE / Greek
            & \yes & \yes &  & & & \yes & &&&\\
        es & Spanish  
            & IE / Italic
            &\yes & \yes &  & & \yes & \yes & &&&\\
        et & Estonian  
            & Uralic 
            & & \yes & &  & & & &&&\\
        eu & Basque  
            & \textit{isolate} 
            & & \yes &  & &  & & &&&\\
        fa & Persian  
            & IE / Indo-Iranian
            & & \yes & &  & & & &&&\\
        fi & Finnish  
            & Uralic 
            & & \yes & &  & & & &&&\\
        fr & French  
            & IE / Italic
            & \yes & \yes & \yes & \yes & \yes & \yes & & &&\\
        ga & Irish  
            & IE / Celtic
            & & & \yes & & & & &&&\\
        he (iw) & Hebrew  
            & Semitic 
            & & \yes &  & & & & &&&\\
        hi & Hindi  
            & IE / Indo-Iranian
            & \yes & \yes & \yes & & & \yes & &&&\\
        hu & Hungarian 
            & Uralic
            & & \yes & &  & & & &&&\\
        id & Indonesian 
            & Austronesian 
            & & \yes & &  & & & &&&\\
        it & Italian  
            & IE / Italic
            & & \yes & &  & \yes & & &&&\\
        ja & Japanese  
            & Japonic 
            & & \yes & &  & & & & \yes&&\\
        jv & Javanese  
            & Austronesian 
            & & \yes & &  & & & &&&\\
        ka & Georgian  
            & Kartvelian 
            & & \yes & \yes &  & & & &&&\\
        kk & Kazakh  
            & Turkic 
            & & \yes & &  & & & &&&\\
        ko & Korean  
            & Koreanic 
            & & \yes & &  & \yes & & &&\yes&\\
        ml & Malayalam 
            & Dravidian 
            & & \yes & &  & & & &&&\\
        mr & Marathi 
            & IE / Indo-Iranian
            & & \yes & &  & & & &&&\\
        nl & Dutch  
            & IE / Germanic
            & & \yes & &  & \yes & & & &&\\
        pt & Portuguese  
            & IE / Italic 
            && \yes & &  & & & &&&\\
        ru & Russian  
            & IE / Balto-Slavic
            & \yes & \yes & &  \yes & & \yes & &&&\\
        sw & Swahili  
            & Bantu 
            & \yes & \yes & \yes & & & \yes & &&&\\
        ta & Tamil  
            & Dravidian
            & & \yes &  & & & & &&&\\
        te & Telugu  
            & Dravidian
            & & \yes & \yes &  & & & &&&\\
        th & Thai  
            & Kra-Dai
            & \yes & \yes & &  & & \yes & &&&\\
        tl (fil) & Tagalog  
            & Austronesian
            & & \yes & \yes &  & & & &&&\\
        tr & Turkish  
            & Turkic
            &\yes & \yes & &  & \yes & \yes & &&&\\
        ur & Urdu  
            & IE / Indo-Iranian
            &\yes & \yes & &  & & \yes & &&&\\
        vi & Vietnamese  
            & Austroasiatic
            &\yes & \yes & &  & & \yes & & &&\\
        zh & Chinese  
            & Sino-Tibetan
            & \yes & \yes & &  \yes & \yes & \yes & &&&\yes\\
        \bhline
    \end{tabular}
    \caption{List of languages used in the experiments.}
    \label{table:list-of-languages}
\end{table*}

\end{document}